\documentclass[10pt,twocolumn,letterpaper]{article}

\usepackage{cvpr}
\usepackage{times}
\usepackage{epsfig}
\usepackage{graphicx}
\usepackage{amsmath}
\usepackage{amssymb}
\usepackage{bm}
\usepackage{subfig}
\usepackage[ruled,vlined]{algorithm2e}
\usepackage{booktabs}
\usepackage{multirow}

\DeclareMathOperator*{\argmin}{arg\,min}

\usepackage[pagebackref=true,breaklinks=true,letterpaper=true,colorlinks,bookmarks=false]{hyperref}

\cvprfinalcopy 

\def\cvprPaperID{5559} 

\ifcvprfinal\fi
\begin{document}

\title{Barycenters of Natural Images -- \\Constrained Wasserstein Barycenters for Image Morphing}

\author{Dror Simon 
\\
Technion\\
Haifa, Israel\\
{\tt\small dror.simon@cs.technion.ac.il}
\and
Aviad Aberdam 
\\
Technion\\
Haifa, Israel\\
{\tt\small aaberdam@cs.technion.ac.il}
}

\maketitle

\begin{abstract}
   Image interpolation, or image morphing, refers to a visual transition between two (or more) input images. For such a transition to look visually appealing, its desirable properties are (i) to be smooth; (ii) to apply the minimal required change in the image; and (iii) to seem ``real'', avoiding unnatural artifacts in each image in the transition. To obtain a smooth and straightforward transition, one may adopt the well-known Wasserstein Barycenter Problem (WBP). While this approach guarantees minimal changes under the Wasserstein metric, the resulting images might seem unnatural. In this work, we propose a novel approach for image morphing that possesses all three desired properties. To this end, we define a constrained variant of the WBP that enforces the intermediate images to satisfy an image prior. We describe an algorithm that solves this problem and demonstrate it using the sparse prior and generative adversarial networks.
\end{abstract}

\section{Introduction}
Image morphing of two input images is a visual effect in which a sequence of images is obtained, transforming one image into the other. By denoting the input images as ${\bm{x}_1,\bm{x}_2\in\mathbb{R}^n}$, the objective is to find a sequence of $N$ images $\{\bm{y}_i\}_{i=1}^N,~\bm{y}_i\in\mathbb{R}^n$ that transform $\bm{x}_1$ to $\bm{x}_2$. 
Generally, there are infinite possible ways of transforming one image into the other. Nevertheless, a pleasant transition should uphold the following properties. First, the difference between any two consecutive frames should be quite similar, leading to a \emph{smooth} steady-paced animation. Second, the overall variation in the entire transition should be \emph{minimal}, avoiding unnecessary changes.

\begin{figure}[t]
    \centering
  \subfloat[Wasserstein barycenters\label{fig:1OT}]{%
       \includegraphics[trim={3px 3px 3px 3px},clip,width=1\linewidth]{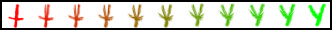}}
  \\
  \subfloat[GAN latent space linear interpolation\label{fig:1GAN}]{%
        \includegraphics[trim={3px 3px 3px 3px},clip,width=1\linewidth]{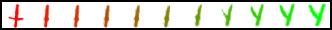}}
  \\
  \subfloat[Ours -- WBP with a GAN as an image prior\label{fig:1ours}]{%
        \includegraphics[trim={3px 3px 3px 3px},clip,width=1\linewidth]{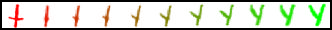}}
  \caption{Morphing a 't' image to a 'y' using 3 different methods, where $\alpha\in\{0,0.1,...,1\}$ (colors are used to emphasize the transition). In Figure \ref{fig:1OT} the intermediate images do not look like English letters. In Figure \ref{fig:1GAN} the rate of the changes varies throughout the transformation. Figure \ref{fig:1ours} demonstrates a smooth transition of English characters. Images taken from the EMNIST dataset \cite{cohen2017emnist}.}
  \label{fig:fig1} 
\end{figure}

\begin{figure*}[t]
  \centering
  \addtocounter{subfigure}{1}
  \subfloat[Wasserstein barycenters\label{fig:2OT}]{%
       \includegraphics[width=0.7\linewidth]{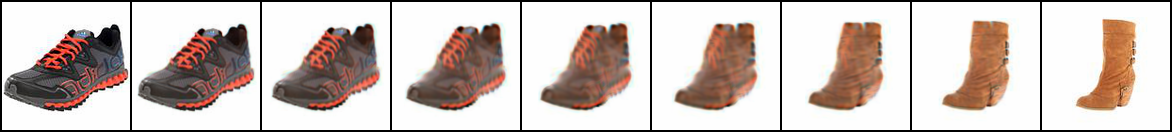}}
  \\
  \addtocounter{subfigure}{-2}
  \subfloat[Image \#1]{%
       \includegraphics[trim={3px 15px 1043px 20px},clip,width=0.12\linewidth]{images/7879163_8072440_OT.png}}
       \addtocounter{subfigure}{1}
      \hfill
  \subfloat[GAN latent space linear interpolation\label{fig:2GAN}]{%
        \includegraphics[trim={0px -13px 0px 0px},clip,width=0.7\linewidth]{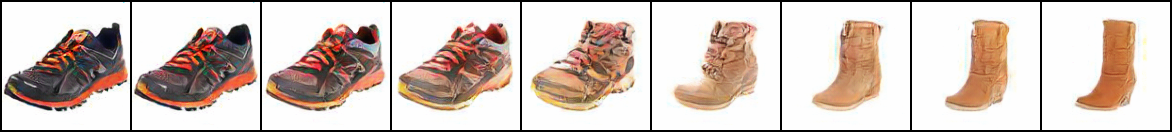}}
       \hfill
  \addtocounter{subfigure}{1}
  \subfloat[Image \#2]{%
       \includegraphics[trim={1043px 20px 3px 15px},clip,width=0.12\linewidth]{images/7879163_8072440_OT.png}}
  \addtocounter{subfigure}{-2}
  \\
  \subfloat[Ours -- WBP with a GAN as an image prior\label{fig:2ours}]{%
        \includegraphics[trim={0px 0 0px 0},clip,width=0.7\linewidth]{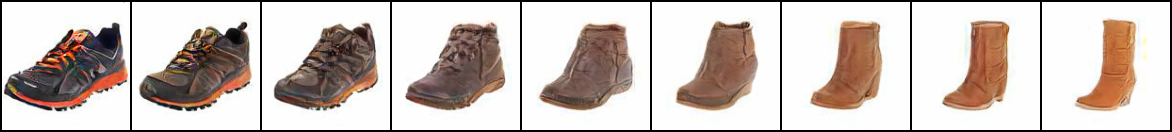}}
  \caption{Morphing a sports shoe to a boot using 3 methods, where $\alpha\in\{0,\frac{1}{8},\frac{2}{8},...,1\}$. In Figure \ref{fig:2OT} the intermediate images look blurry and unrealistic. In Figure \ref{fig:2GAN} at first the shoe hardly changes and then immediately changes to a boot. Figure \ref{fig:2ours} demonstrates a smooth transition in both color and shape. Images taken from the Zappos50k dataset \cite{yu2014fine}.}
  \label{fig:fig2} 
\end{figure*}

The naive solution to consider for image morphing is a simple linear interpolation between the two images, i.e. ${\bm{y}_i=\frac{N+1-i}{N+1}\bm{x}_1 + \frac{i}{N+1}\bm{x}_2}$. While this method indeed produces a smooth transition, it leads to unnatural intermediate samples that contain unpleasant double-exposure artifacts. Therefore, to obtain a pleasant transition, an additional requirement is needed.

An approach that overcomes the double-exposure artifact, is solving the Wasserstein Barycenter Problem (WBP) \cite{cuturi2014fast,cuturi2016smoothed}. The Wasserstein barycenter is the probability distribution function that minimizes the mean of its Wasserstein distances \cite{ruschendorf1985wasserstein} to each element in a given set of probability distributions. Considering two input probability distributions located on the simplex $\{\bm{p}_1,\bm{p}_2\}\in\Sigma_n$ the WBP is then defined as
\begin{equation}
    \bm{p}_{\alpha} = \argmin_{\bm{q}\in\Sigma_n}~ (1-\alpha)\mathcal{W}_2^2(\bm{p}_1,\bm{q}) + \alpha\mathcal{W}_2^2(\bm{p}_2,\bm{q}),
    \label{eq:barycenter}
\end{equation}
where $\alpha\in[0,1]$ and $\mathcal{W}_2(\bm{p},\bm{q})$ denotes the $\ell_2$ Euclidean Wasserstein distance between $\bm{p}$ and $\bm{q}$ (see Section \ref{sec:WBP}).
To obtain a sequence that morphs the distribution $\bm{p}_1$ to $\bm{p}_2$ smoothly, a common approach is to solve Equation \eqref{eq:barycenter} for a linear series of $\alpha$ values, e.g. $\alpha \in\frac{1}{N+1}\left\{1,2,...,N\right\}$. Indeed, solving the WBP for two input images, leads to a smooth (regular) and direct transition while avoiding ghosting artifacts.\footnote{This usually requires a pre-processing normalization step.} That said, the intermediate samples do not necessarily seem ``natural'' as can be seen in Figures \ref{fig:1OT} and \ref{fig:2OT}. To overcome this issue, one may replace the $\ell_2$ Euclidean metric with the geodesic distance over the manifold of natural images. However, this manifold is typically unknown or very complex, making this approach impractical.

In order to obtain natural intermediate images, recent works have suggested the use of Generative Adversarial Networks (GAN) \cite{goodfellow2014generative,radford2015unsupervised,isola2017image,zhu2016generative}. In this architecture, a generative network $G(\cdot)$ maps vectors $\bm{z}_i\in\mathbb{R}^m,~m<n$ from a low dimensional latent space to high dimensional images. When two images and their matching latent representations are given, i.e. $G(\bm{z}_i)=\bm{x}_i$, a transition is obtained by linearly interpolating the two latent vectors as follows:
\begin{equation}
    \bm{y}_i = G\left((1-\alpha_i)\bm{z}_1 + \alpha_i\bm{z}_2\right),
\end{equation}
with $\alpha_i=\frac{i}{N+1}$. Since each interpolated image $\bm{y}_i$ is an output of the generative network, each image follows an image prior, leading to a natural-looking transition. However, as we show in this work, these transitions do not necessarily obey the desired properties mentioned earlier. First, the pace of the changes might vary throughout the transformation as demonstrated in Figures \ref{fig:1GAN} and \ref{fig:2GAN}, where most of the transition is concentrated in one or two frames. Second, the change itself might not be direct and minimal. For example, in Figure \ref{fig:2GAN} the colors become too bright before darkening back again.

The main contribution of this work is in providing a novel algorithm for solving a constrained form of the WBP. Furthermore, in this work we introduce a novel approach for image morphing that is based on the Euclidean WBP but with additional constraints on the obtained intermediate images. Concretely, we enforce each of the images in the sequence to reside on the manifold of natural images by using image priors, leading to a transition that fulfills all of the aforementioned requirements. Moreover, we present an approach to measure these three properties numerically and show the advantage of our method.

\section{Previous Work}
\label{sec:previous}
Image morphing has been studied and evolved for over three decades. Classical methods \cite{wolberg1998image} have relied on a simple cross-dissolve operation together with a geometric warp of two images, using a dense correspondence map, which is typically hard to obtain automatically. In some cases, however, manual correspondence maps were avoided. For example, in \cite{zhu2007image}, a method that is based on optimal-transport was suggested to obtain a short time domain interpolation, e.g. interpolating two consecutive video-frames. A recent work has suggested a morphing process in which the intermediate images are generated by patches of the input ones, constraining their similarity \cite{shechtman2010regenerative}. This method does not require such maps even when the input images differ substantially. Later, \cite{zhu2016generative} has extended this concept by generalizing the local patch-based constraints to a single global one. To morph one image into the other, the authors have suggested to traverse the manifold of natural images. To this end, they first project the input images onto the latent space of a trained GAN, and then linearly interpolate these latent vectors. This transformation is used to compute a motion and color flow, which is then applied on one of the inputs. Hence, the final transformation is actually a geometrical warp of the input image, as opposed to being generated by the model. In this work, we further extend this approach by traversing the latent space in a non-linear manner. This path is obtained by solving the Wasserstein barycenter problem for the input images. Moreover, we discard the use of the flow fields by increasing the resolution of the generated images. Furthermore, our method is not restricted to GANs and can be used with any image prior.

\section{The Wasserstein Barycenter Problem}
\label{sec:WBP}
Before describing the WBP, we first provide a brief overview on optimal transport and Wasserstein distances. For an in-depth review of the topic, the reader is referred to \cite{peyre2019computational} and \cite{villani2008optimal}.

\subsection{Symbols and Notations}
We define $\mathcal{D}$ as a space created by regular samples in $\mathbb{R}^2$, leading to an $n_1\times n_2$ grid of pixels, where $n_1 n_2=n$. For simplicity, we refer to $\bm{p}\in\mathcal{D}$ as a 1-dimensional vector of size $n$. The symbol $\mathcal{D}_+^1$ denotes the space of probability measures defined on $\mathcal{D}$, i.e. if $\bm{p}\in\mathcal{D}_+^1$, then $\sum_{i=1}^n p_i = 1$ and $p_i\geq 0$ where the element $p_i$ is mapped to the $i$-th pixel in $\mathcal{D}$. Finally, we use $d_{\mathcal{D}}(i,j)$ to denote the Euclidean distance between pixels $i$ and $j$ in the grid defined by $\mathcal{D}$.

\subsection{Optimal Transport}
Given a source and target distributions $\bm{p}_1,\bm{p}_2\in\mathcal{D}_+^1$, it is possible to transform one to the other using a transportation-plan $\bm{P}\in\mathbb{R}_+^{n\times n}$. This transportation plan describes the amount of mass to be passed from each pixel in $\bm{p}_1$ to each pixel in $\bm{p}_2$, while preserving mass conservation rules. The set containing all possible plans $U(\bm{p}_1,\bm{p}_1)$ is defined as:
\begin{equation}
    U(\bm{p}_1,\bm{p}_1) = \left\{ \bm{P}\in\mathbb{R}_+^{n\times n} \middle | \bm{P}\cdot\mathbf{1}_n=\bm{p}_1 \cap \bm{P}^T\cdot\mathbf{1}_n=\bm{p}_2 \right\},
\end{equation}
where $\mathbf{1}_n$ is an all-ones vector of size $n$. For a given cost matrix $\bm{C}\in\mathbb{R}^{n\times n}$, \emph{optimal transport} is defined as the transportation plan $\bm{P}^*$ which is the minimizer of:
\begin{equation}
    \mathcal{L}_C(\bm{p}_1,\bm{p}_2) = \min_{\bm{P}\in U(\bm{p}_1,\bm{p}_2)} \sum_{i=1}^n \sum_{j=1}^n \bm{P}_{i,j}\bm{C}_{i,j}.
    \label{eq:ot}
\end{equation}
Specifically, when the matrix $\bm{C}$ is a distance matrix, then $\mathcal{L}_C(\bm{p}_1,\bm{p}_2)$ is referred to as a \emph{Wasserstein distance}. For example, when the Euclidean $\ell_2$ distance is used, Eq. \eqref{eq:ot} is equivalent to:
\begin{equation}
    \mathcal{L}_{d_{\mathcal{D}}}(\bm{p}_1,\bm{p}_2) = \min_{\bm{P}\in U(\bm{p}_1,\bm{p}_2)} \sum_{i=1}^n \sum_{j=1}^n \bm{P}_{i,j}d_{\mathcal{D}}(i,j),
    \label{eq:wasserstein_dist}
\end{equation}
and we denote $\mathcal{W}_2(\bm{p}_1,\bm{p}_2) \triangleq \sqrt{\mathcal{L}_{d_{\mathcal{D}}}(\bm{p}_1,\bm{p}_2)}$. Indeed, as the name suggests, the Wasserstein distance is also a distance metric.

To find the minimizer $\bm{P}^*$ of Eq. \eqref{eq:wasserstein_dist}, one needs to solve a LP problem of size $n\times n$. For example, an image of size $128\times 128$ leads to a LP of size $16,384^2\approx 2\times 10^8$, making it an impractical task. To overcome this issue, we seek to approximate problem \eqref{eq:wasserstein_dist}. A common approximation is the use of an entropic regularization \cite{cuturi2013sinkhorn}:
\begin{align}
    &\mathcal{W}_2^2(\bm{p}_1,\bm{p}_2) = \min_{\bm{P}\in U(\bm{p}_1,\bm{p}_2)} \sum_{i=1}^n \sum_{j=1}^n \bm{P}_{i,j}d_{\mathcal{D}}(i,j) - \epsilon \mathbf{H}(\bm{P}),
    \label{eq:wasserstein_dist_entropic}
    \\
    & \mathbf{H}(\bm{P}) \triangleq -\sum_{i, j} \bm{P}_{i, j}\left(\log \left(\bm{P}_{i, j}\right)-1\right).
\end{align}
This regularization stabilizes the solution by making the problem strictly convex and the solution can be found efficiently using the Sinkhorn algorithm \cite{sinkhorn1967diagonal}. Hereinafter, we denote $\mathcal{W}_2$ as the entropic-regularized Wasserstein distance.

\subsection{Wasserstein Barycenters}
For any given distance metric $d$, the \emph{barycenter} of a set of inputs $\left\{\bm{x}_i\right\}_{i=1}^n$ and corresponding weights $\left\{w_i\right\}_{i=1}^n$ where $w_i\geq 0$ and $\sum_i w_i = 1$ is defined as:
\begin{equation}
    \bm{x}_{\text{barycenter}} = \argmin_{\bm{x}} \sum_{i=1}^n w_i d(\bm{x},\bm{x}_i)^p,
\end{equation}
where $p \geq 1$.
Specifically, the Wasserstein barycenter problem is defined as the probability measure that minimizes the sum of $p$-powered Wasserstein distances to a set of probability measures $\{\bm{p}_i\}_{i=1}^n$:
\begin{equation}
    \bm{q}_{\text{barycenter}} = \argmin_{\bm{q}\in\mathcal{D}_+^1} \sum_{i=1}^n w_i \mathcal{W}_2^2(\bm{q},\bm{p}_i),
    \label{eq:wbarycenter}
\end{equation}
where in \eqref{eq:wbarycenter} we chose $p=2$. This problem is strictly convex and various efficient solvers have been suggested \cite{cuturi2014fast, cuturi2016smoothed, solomon2015convolutional, bonneel2016wasserstein}. Wasserstein barycenters have been used for various applications in image processing and shape analysis, including texture mixing \cite{rabin2011wasserstein}, color transfer \cite{ferradans2014regularized, solomon2015convolutional}  and shape interpolation \cite{solomon2015convolutional}. In the following section, we propose a novel solution for a constrained version of the Wasserstein barycenter problem, and use it to obtain a natural-looking barycenter of images.

\section{The Proposed Approach}
\begin{figure}[t]
    \centering
    \includegraphics[trim={0 200 0 0},clip,width=1\linewidth]{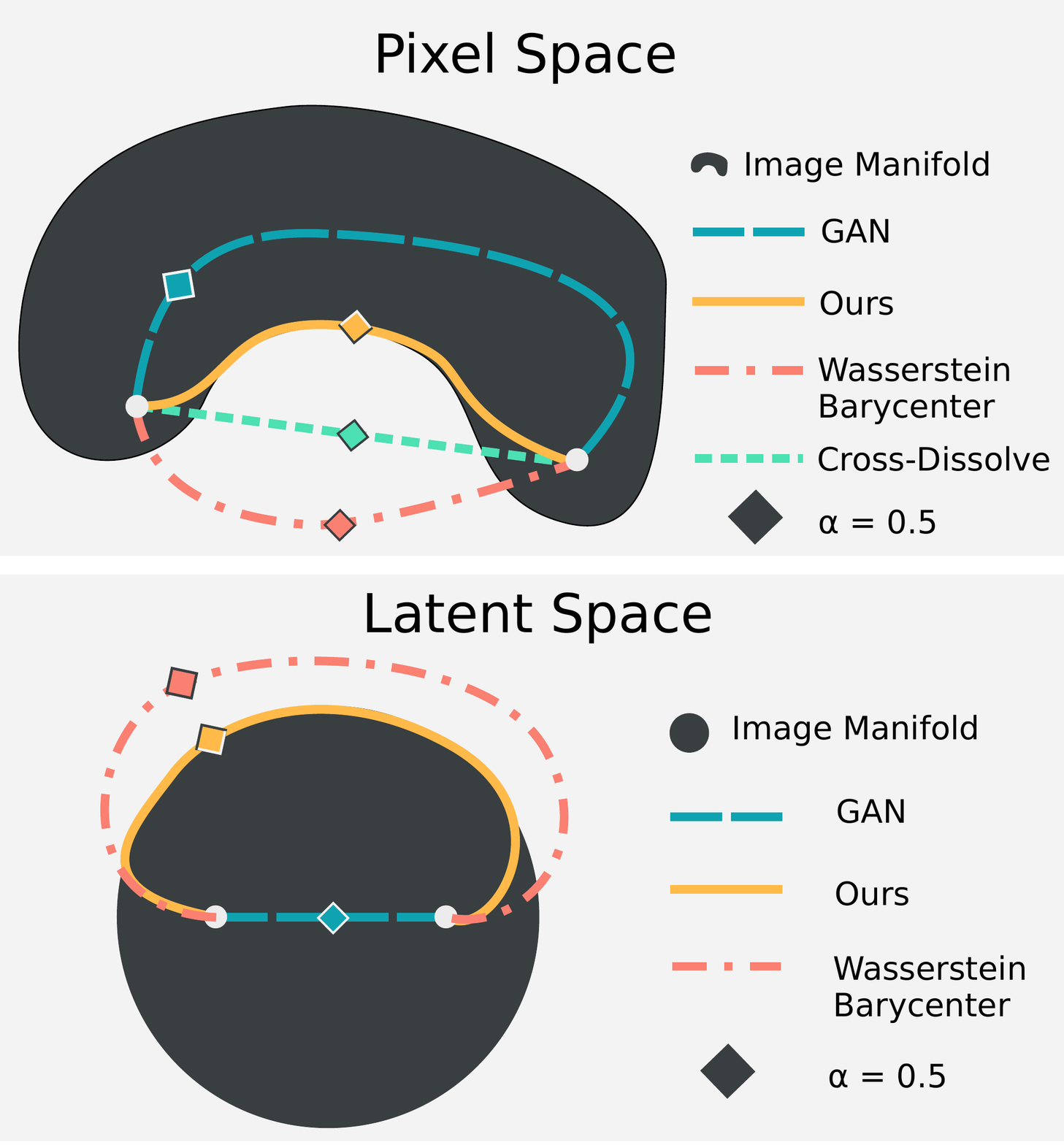}
    \caption{An illustration of the morphing process of an image, setting $\alpha=0 \to 1$, using several approaches in the pixel space and the latent space of a trained GAN.}
    \label{fig:drawing}
\end{figure}

To morph one image to the other, while obtaining natural looking intermediate images, we suggest to restrict the obtained images to satisfy some prior. Formally, we add a constraint to the barycenter problem (Eq. \eqref{eq:barycenter}) that limits the result to lie on a manifold $\mathcal{M}$:
\begin{equation}
    \bm{p}_{\alpha} = \begin{cases}\argmin_{\bm{q}\in\Sigma_n}~ (1-\alpha)\mathcal{W}_2^2(\bm{p}_1,\bm{q}) + \alpha\mathcal{W}_2^2(\bm{p}_2,\bm{q})\\
    \text{s.t.}\quad \bm{q}\in \mathcal{M},
    \end{cases}
    \label{eq:CWBP}
\end{equation}
As this problem might be hard to solve directly, we shall place an auxiliary variable $\bm{r}=\bm{q}$:
\begin{equation}
    \bm{p}_{\alpha} = \begin{cases}\argmin\limits_{\bm{q}\in\Sigma_n, r}~ (1-\alpha)\mathcal{W}_2^2(\bm{p}_1,\bm{q}) + \alpha\mathcal{W}_2^2(\bm{p}_2,\bm{q})\\
    \text{s.t.}\quad \bm{r}\in \mathcal{M}, ~ \bm{r}=\bm{q}.
    \end{cases}
\end{equation}
The Augmented Lagrangian of this problem is
\begin{equation}
    \bm{p}_{\alpha} = \begin{cases}\argmin\limits_{\bm{q}\in\Sigma_n, \bm{r}, \bm{u}}~ (1-\alpha)\mathcal{W}_2^2(\bm{p}_1,\bm{q}) + \alpha\mathcal{W}_2^2(\bm{p}_2,\bm{q})\\
    \quad\quad\quad\quad+ \frac{\mu}{2}\|\bm{r}-\bm{q}+\bm{u}\|_2^2\\
    \text{s.t.}\quad \bm{r}\in \mathcal{M},
    \end{cases}
    \label{eq:augmented}
\end{equation}
where $\mu>0$.
This optimization problem can be solved using the Alternating Direction Method of Multipliers (ADMM) \cite{boyd2011distributed}, leading to the following steps (see Algorithm \ref{algorithm:admm}). First, we find a solution $\bm{q}$ to a regularized version of the WBP. This problem is strictly convex and has been previously studied. In our work we follow \cite{cuturi2016smoothed} which proposes a descent algorithm on the dual problem. The second step is a projection of the previous step result $\bm{q}$ onto the manifold $\mathcal{M}$. The third and the final step is a simple update of the dual variable $\bm{u}$. These steps are repeated until convergence is achieved. Figure \ref{fig:drawing} illustrates the differences between our approach and other image morphing approaches, specifically when using a GAN as an image prior.

\begin{algorithm}[t]
\SetAlgoLined
\SetKwInOut{Input}{input}\SetKwInOut{Output}{output}\SetKwInOut{Init}{initialize}
\Input{Input densities $\{p_1,p_2\}$, an initial guess $\bm{q}^0$ and a threshold $\epsilon$}
\Output{Constrained barycenter $\bm{q}$.}
\BlankLine
\Init{$\bm{u}\gets \bm{0}$, $k\gets 0$, $\bm{r}^0 \gets \bm{q}^0$}
    \Repeat{$\|\bm{q}^k-\bm{r}^k\|<\epsilon$}{
        \begin{flalign*}
            \bm{q}^{k+1} \gets &\argmin_{\bm{q}\in\Sigma_n} \alpha\mathcal{W}_2^2(\bm{p_1},\bm{q}) + (1-\alpha)\mathcal{W}_2^2(\bm{p_2},\bm{q})&\\
            &\quad\quad + \frac{\mu}{2}\|\bm{r}^k-\bm{q}+\bm{u}^k\|_2^2&
        \end{flalign*}
        \begin{flalign*}
            \bm{r}^{k+1}\gets &\argmin_{\bm{r}\in \mathcal{M}} \frac{\mu}{2}\|\bm{r}-\bm{q}^{k+1}+\bm{u}^k\|_2^2 &
        \end{flalign*}
        \begin{flalign*}
            &\bm{u}^{k+1}\gets \bm{u}^k + \bm{r}^{k+1} - \bm{q}^{k+1}&
        \end{flalign*}
        $k\gets k+1$
    }
    \Return{$\bm{q}^{k}$}
    \caption{Constrained Wasserstein Barycenters}
    \label{algorithm:admm}
\end{algorithm}

In cases where the manifold $\mathcal{M}$ is convex, the optimization problem \eqref{eq:augmented} is convex, and convergence to a global minimum is guaranteed. That said, manifolds of interest, such as those of natural images, are often not convex (otherwise a simple linear interpolation between images would suffice) and therefore, only a local minimum is not guaranteed. Nevertheless, as we show in section \ref{sec:experiments}, the obtained results are visually appealing. Note that this approach can be applied for a variety of priors, affecting only the second step in Algorithm \ref{algorithm:admm}. In the following subsections we demonstrate our method on the sparse prior and on GANs.

\subsection{Sparse Prior}
\label{sec:sparse}
A well-known prior for various signal processing tasks is the sparse representation prior \cite{elad2010sparse,bruckstein2009sparse,elad2006image}. This model assumes that a signal $\bm{x}\in\mathbb{R}^n$ is constructed by a linear combination of only a few columns, also referred to as \emph{atoms}, taken from a fixed matrix $\bm{D}\in\mathbb{R}^{n\times m}$, known as a \emph{dictionary}. When a signal $\bm{y}$ is given, projecting it onto the model consists of finding its sparse representation vector $\bm{\alpha}$:
\begin{equation}
    \hat{\bm{\alpha}} = \argmin_{\bm{\alpha}} \|\bm{y}-\bm{D}\bm{\alpha}\|_2^2 ~\text{s.t.}~ \|\bm{\alpha}\|_0< k,
    \label{eq:sparse}
\end{equation}
for some $k\in\mathbb{N}$, typically much smaller than $m$. Generally, Eq. \eqref{eq:sparse} is NP-hard \cite{natarajan1995sparse} and various approximation algorithms have been suggested to solve this problem, such as the Orthogonal Matching Pursuit (OMP) and the Basis Pursuit (BP) algorithms \cite{pati1993orthogonal,chen1994basis}. Once the representation vector $\hat{\bm{\alpha}}$ is found, the reconstructed signal is simply $\bm{y}^{\text{proj}} = \bm{D}\hat{\bm{\alpha}}$.

In our approach, constraining the resulting barycenter image to satisfy the sparse representation prior, changes the second step in Algorithm \ref{algorithm:admm} to a sparse coding algorithm, e.g. OMP. In our experiments, we further improve the visual results by approximating the MMSE estimator of $\hat{\bm{\alpha}}$ using stochastic resonance \cite{simon2019mmse}.

\subsection{Generative Adversarial Networks}
\label{sec:GAN}
In the GAN setting \cite{goodfellow2014generative,radford2015unsupervised}, a generative network $G(\bm{z}): \mathbb{R}^m\to\mathbb{R}^n$ and a discriminative one $D(\bm{y}): \mathbb{R}^n\to\{0,1\}$ contest against each other. Given a dataset, the former is trained to generate samples from it when given a random input vector $\bm{z}$, while the latter is trained to distinguish the generated data from the original one. This approach leads to a model that is able to generate new data samples with statistical properties that are similar to the training set, by sampling random vectors $\bm{z}$ from the latent space of the model, and passing them through $G$. 

In order to use the generative network for image morphing, an inverse mapping, i.e. a mapping from an input image $\bm{x}$ to its latent representation vector $\bm{z}$, is required. To obtain this mapping, we follow the approach described in \cite{zhu2016generative} that we now briefly describe here. Once the generative network is trained, we train an encoder $E(\bm{x}):\mathbb{R}^n\to\mathbb{R}^m$, such that $G(E(\bm{x}))$ is similar to the input $\bm{x}$:
\begin{equation}
    E(\bm{x}) = \bm{z}^* = \min_{\bm{z}} \mathcal{L}\left(\bm{x},G(\bm{z(\bm{x})})\right),
\end{equation}
where $\mathcal{L}$ is a pixel-wise $\ell_2$ loss in simple cases such as MNIST \cite{lecun2010mnist}. For more complex images such as Zappos50k \cite{yu2014fine}, the loss $\mathcal{L}$ is extended to a weighted sum of pixel-wise $\ell_2$ and features extracted from AlexNet \cite{krizhevsky2012imagenet} trained on ImageNet \cite{deng2009imagenet}. This encoder-decoder scheme $G\circ E:\mathbb{R}^n\to\mathbb{R}^n$ may be perceived as a projection of the input signal onto the manifold of natural images. Therefore, To use GAN as a prior in our approach, the second step in Algorithm \ref{algorithm:admm} is implemented by a simple feedforward activation of the obtained encoder-decoder.

\section{Quantifying the Desired Properties}
\label{sec:quantify}
Above, we described 3 desired attributes for a natural looking image transformation: (i) to be smooth (regular), i.e. change at a constant pace; (ii) to be as minimal and direct as possible, avoiding unnecessary changes; and (iii) to include natural-looking images. To quantitatively show the advantage of our approach over other alternatives, we propose to measure each of these attributes as follows:
\begin{enumerate}
    \item Regularity -- to evaluate the smoothness of a transition, we propose to measure the distance between every two consecutive frames, and then compute the standard deviation of these distances over the entire transition. A steady paced transition results in a very low standard deviation, whereas irregular changes in the transformation correspond to a high variance. Since the Euclidean norm does not fit to measure movements of pixels in the image \cite{wang2009mean}, we adopt the Wasserstein distance for this task as it evaluates the minimal effort required to transport each pixel from one frame to the other.
    \item Minimal -- a minimal transition consists of a small number of pixel movements during the transformation process. As before, we adopt the Wasserstein distance for this task as it is a natural metric to quantify these movements. To evaluate the cost of the entire transition, we propose to average the Wasserstein distances between every two successive frames in the transformation.
    \item Natural looking images -- to evaluate the affinity of an image to the class of natural images, we first train an autoencoder on a training set drawn from the chosen dataset. Once this model is trained, we feed each of the images generated in the transformation through the model, and compute their $\ell_2$ distance to their own projection on the manifold characterized by the autoencoder.
\end{enumerate}

\section{Experiments}
\label{sec:experiments}

\subsection{MNIST}
\begin{figure}[t]
    \centering
    \includegraphics[width=1\linewidth]{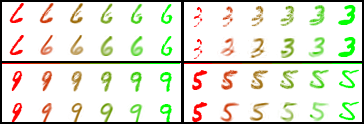}
    \caption{Wasserstein barycenters and our approach using a sparse prior in rows 1 and 2 of every subfigure respectively.}
  \label{fig:sparse} 
\end{figure}

We first demonstrate our method using the sparse representation model. From our experiments, the generative capabilities of this model seem inferior to those of newer alternatives such as GANs. Nevertheless, this example exposes the generality of our approach regarding the chosen prior. We start by learning a dictionary for the training set of the MNIST dataset \cite{lecun2010mnist}, using online dictionary learning \cite{mairal2010online}. Then, we randomly select two test images of the same digit and morph one to the other using Algorithm \ref{algorithm:admm}, as described in Section \ref{sec:sparse}. For comparison, we show the results of the morphing process using the unconstrained Wasserstein barycenters. As demonstrated in Figure \ref{fig:sparse}, constraining the barycenter outcomes to satisfy the sparse prior yields sharper images that look like real digits, whereas the Wasserstein barycenter approach has no such guarantee.

\begin{figure}[t]
    \hspace{0.06\linewidth} Wasserstein \hspace{0.17\linewidth} Ours \hspace{0.2\linewidth} GAN \\
    \phantom{.}\hspace{0.055\linewidth} barycenters
    \\
    \includegraphics[width=0.325\linewidth]{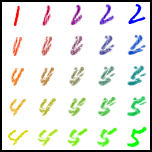}
    \includegraphics[width=0.325\linewidth]{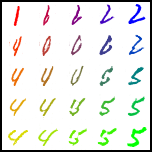}
    \includegraphics[width=0.325\linewidth]{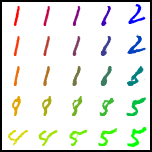}
    \\
    \includegraphics[width=0.325\linewidth]{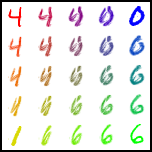}
    \includegraphics[width=0.325\linewidth]{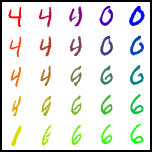}
    \includegraphics[width=0.325\linewidth]{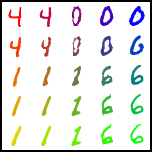}
    \\
    \includegraphics[width=0.325\linewidth]{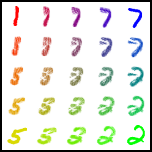}
    \includegraphics[width=0.325\linewidth]{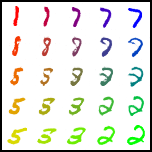}
    \includegraphics[width=0.325\linewidth]{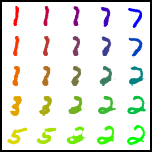}
    \caption{Barycenters of 4 input images, where each one is placed in a corner of the square. We compare 3 methods: Wasserstein barycenters, Algorithm \ref{algorithm:admm} using DCGAN as an image prior, and DCGAN latent space bilinear interpolation. Colors are used to emphasize the interpolation.}
  \label{fig:mnist_gan_square} 
\end{figure}

We continue our experiments with employing GAN as an image prior, as described in Section \ref{sec:GAN}. Specifically, we use the DCGAN architecture \cite{radford2015unsupervised}. This prior is much more potent and is able to generate digit-like images, even when transforming between different digits. In this case, we experiment with barycenters of 4 input images. To do so, we modify Eq. \eqref{eq:barycenter} to include a convex combination of 4 Wasserstein distances, one from each source image. Figure \ref{fig:mnist_gan_square} presents a comparison between our approach, the standard Wasserstein barycenters, and a bilinear interpolation of the latent vectors in the GAN setting. In contrast to the Wasserstein barycenters results, both our method and latent space bilinear interpolation produce natural digits. However, in the GAN setting, the pace is not consistent, leading to false barycenters. For example, in the first row, the image in the center does not look like an ``average'' of all the others, but is rather similar to the digit ``1'', inserted at the top-left corner.

\subsection{Extended MNIST}
\begin{figure*}[t]
    \centering
    \includegraphics[width=1\linewidth]{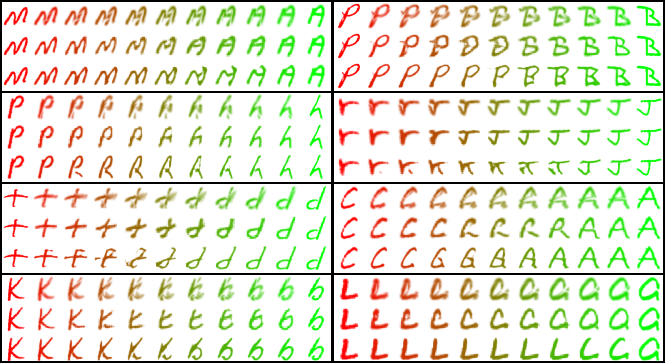}
    \caption{Morphing English character images using: Wasserstein barycenters, our approach using DCGAN as an image prior and DCGAN latent space linear interpolation in the $1$-st, $2$-nd and $3$-rd row of every subfigure respectively.}
  \label{fig:emnist} 
\end{figure*}

The Extended-MNIST dataset \cite{cohen2017emnist} contains English characters that are more complicated than digits, and using the sparse prior leads to unpleasant results. Therefore, to obtain natural looking images, we focus our experiment in the DCGAN setting. Figures \ref{fig:fig1} and \ref{fig:emnist} demonstrate transitions using Wasserstein barycenters, linear interpolation in the latent space, and our method employing the DCGAN as the image prior. As before, it can be observed that the morphs obtained by the latent space interpolation do not result in a steady paced transition. At the bottom right example, the `L' character hardly changes over the entire morphing process and most of the transformation occurs at the last 2 steps, i.e. $\alpha \in [0.8,1]$. Furthermore, in the second example from the top on the right-hand side, it seems that transition from an `r' to a `J' in the latent interpolation case is not as straightforward as in our approach. Regarding the Wasserstein barycenter, the outcome images are blurry and often do not look like real English characters.

In addition to the provided visual results, Table \ref{table:quantify} presents the averaged evaluation of all three methods, using the metrics specified in Section \ref{sec:quantify}, on 1500 randomly chosen image pairs. These results show the advantage of our method as it obtained a straightforward and steadier paced transition compared to a latent space linear interpolation, while still being close to the desired manifold, as opposed to the Wasserstein barycenters approach.

\begin{table}[t]
\begin{tabular}{@{}lccc@{}}
\toprule
Method      & Regularity             & Total Dist.         & Dist. to Manifold                            \\ \midrule
Wasserstein & \multirow{2}{*}{0.033} & \multirow{2}{*}{10.95} & \multirow{2}{*}{$4.96\times 10^{-4}$} \\
Barycenters &                        &                        &                                                 \\
DCGAN         & 0.983                  & 17.48                  & $2.35\times 10^{-4}$                                               \\
Ours        & 0.232                  & 12.58                  & $2.37\times 10^{-4}$                                               \\ \bottomrule
\end{tabular}
\caption{The averaged quantified properties of 1500 randomly chosen transformations from the EMNIST dataset. In all parameters lower is better.}
\label{table:quantify}
\end{table}

\subsection{Shoe Images -- UT Zappos50K}
The Zappos50K \cite{yu2014fine} is a much more complicated dataset compared to the previous two. Specifically, it contains more details and higher resolution. To train a GAN capable of generating such images, we split the training process in two, somewhat similar to the training scheme described in StackGAN \cite{zhang2017stackgan}. First, we downscale the images to $64\times 64$, and train a DCGAN model, as well as an encoder, as described in Section \ref{sec:GAN}. The output images of this model are very smooth and lack fine high-frequency details. To add these details, we train an additional generative model. To this end, we generate a dataset of input-output image pairs as follows: the input images are the output of the encoder-decoder scheme, upsampled to $128\times 128$, whereas the output images are the original ones downscaled to $128\times 128$. This dataset is used as a training set to a pix2pix model \cite{isola2017image}. To summarize, our projection scheme consists of the following stages: (i) project the input image to the DCGAN's latent space using a trained encoder; (ii) generate a low-frequency $64\times 64$ image using a DCGAN; (iii) upscale the image to $128\times 128$; and (iv) feed the image into a pix2pix model to generate high frequency details.

\begin{figure*}[t]
    \centering
    \includegraphics[width=0.88\linewidth]{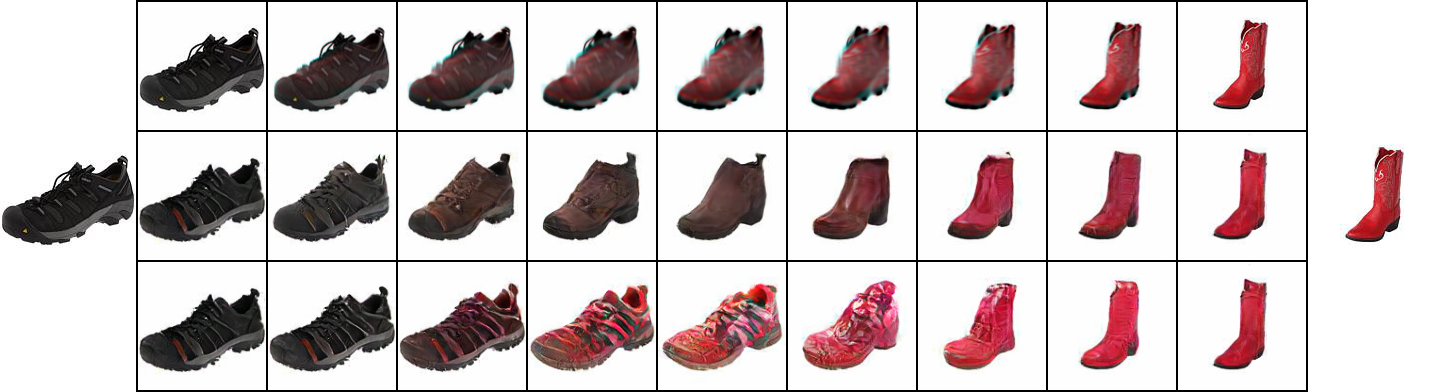}
    \includegraphics[width=0.88\linewidth]{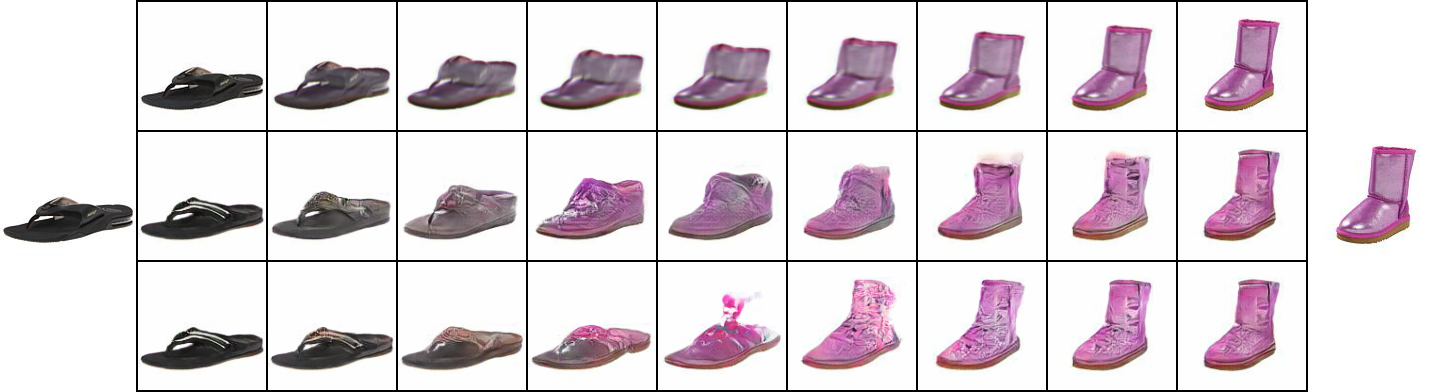}
    \includegraphics[width=0.88\linewidth]{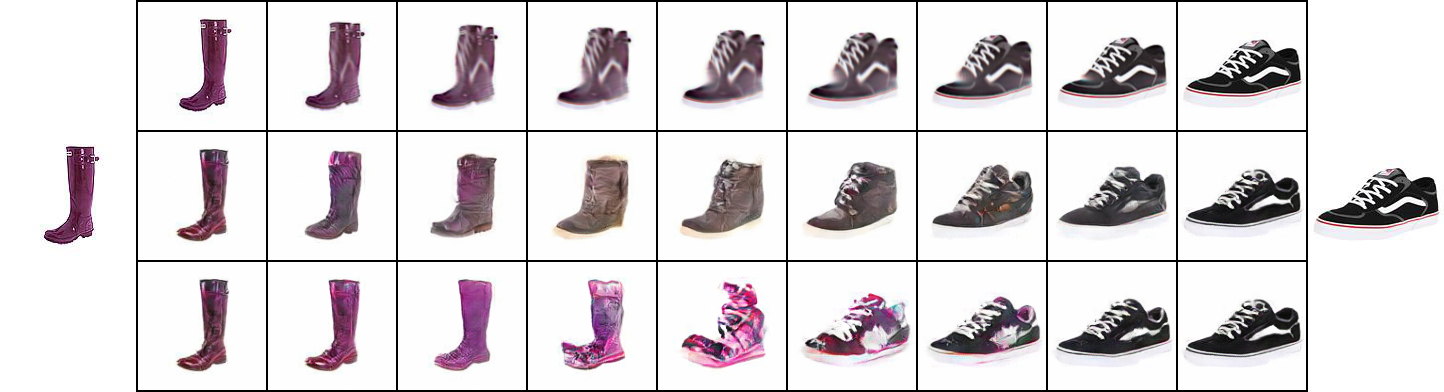}
    \includegraphics[width=0.88\linewidth]{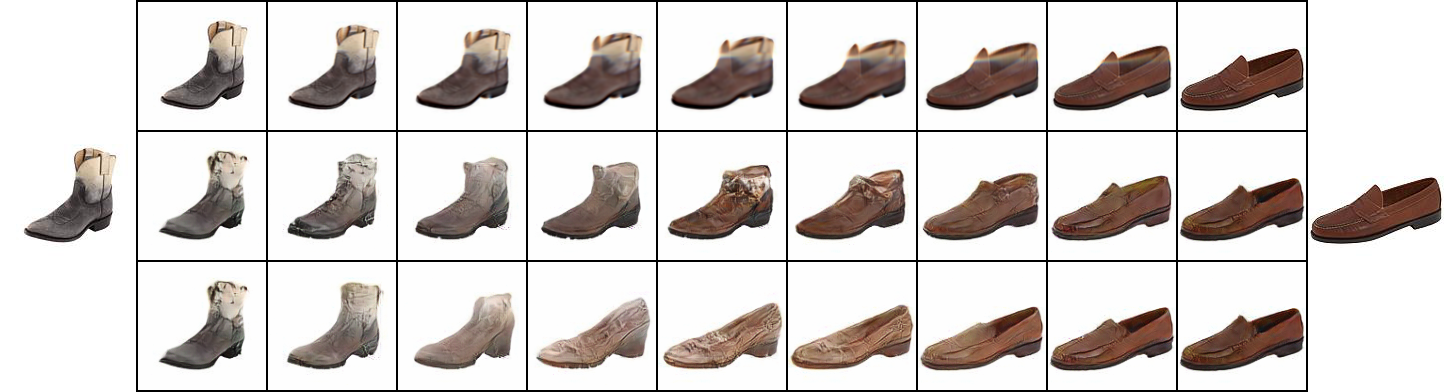}
    \includegraphics[width=0.88\linewidth]{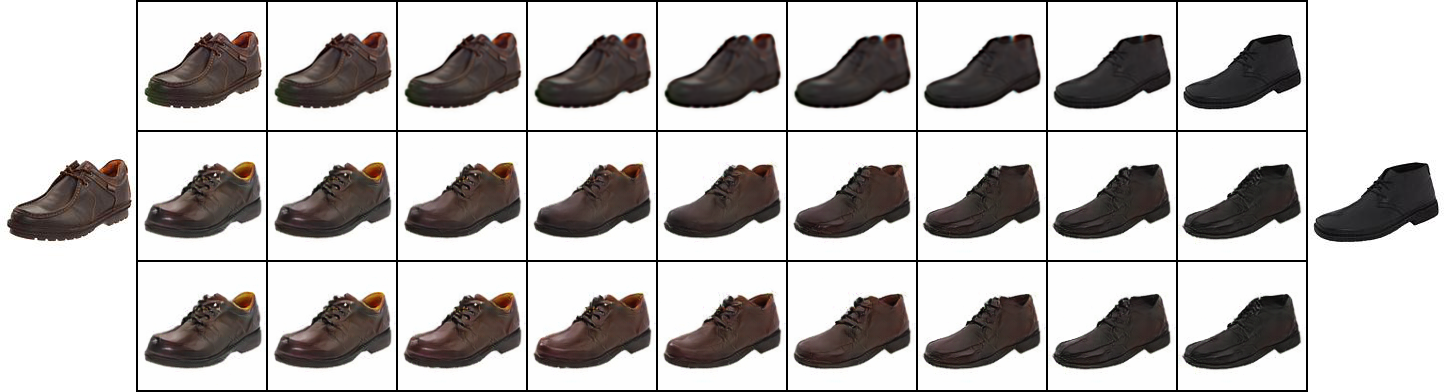}
    \caption{Morphing shoe images using: Wasserstein barycenters, our approach using GANs as an image prior and GAN latent space linear interpolation ($1$-st, $2$-nd and $3$-rd row of every subfigure respectively).}
  \label{fig:shoes} 
\end{figure*}

Once the models are trained, we compare the three following methods. The first is a standard Wasserstein barycenter solution (applied on each color channel separately). The second approach is our proposed algorithm, i.e. project each of the Wasserstein barycenters to the manifold of natural images, using the trained generative models, and the third is the common transition achieved using GANs as follows. Each input image is projected onto the latent space, using the encoder. Then, to obtain a transition, we linearly interpolate the two latent vectors and pass the interpolated vectors through the DCGAN and pix2pix models. From our experiments, iterating once produces the best results. The results of our experiments are presented in Figures \ref{fig:fig2} and \ref{fig:shoes}. Both our method and the GAN alternative provide natural images most of the time. Furthermore, in cases where the two input images are similar in shape and color the difference between the two approaches seems mild (see the last example in Figure \ref{fig:shoes}). However, when the contour or the hue of the two input images differ significantly, our approach brings on a much more steady and straightforward transition to both the shape and the colors of the images. 

\section{Conclusions}
In this work we introduced a novel solution to a constrained variant of the well-known Wasserstein barycenter problem. While our algorithm is general, we propose to use it to obtain a natural barycenter (average) of two or more input images, which can be used to generate a smooth transition from one to the other. For this purpose, we suggest constraining the barycenter to an image prior. Specifically, we demonstrate our approach using the sparse prior and generative adversarial networks. We compare our method with the unconstrained variant of the WBP and a linear interpolation of latent vectors of GANs and show the advantage of the former in terms of the smoothness of the transition, the minimal quantity of changes, and the natural look of the acquired images both visually and numerically. Moreover, we believe our approach of solving the WBP in its constrained form can be used in a variety of applications other than image morphing, e.g. pitch interpolation of two speakers, image style transfer and more, and we will focus our future work on such extensions.

\clearpage
{\small
\bibliographystyle{ieeetr}
\bibliography{egbib}
}

\end{document}